# A Computer Vision Application for Assessing Facial Acne Severity from Selfie Images

Tingting Zhao
Microsoft Corporation
Redmond, WA
tingting.zhao@microsoft.com

Hang Zhang
Microsoft Corporation
Redmond, WA
hangzh@microsoft.com

Jacob Spoelstra
Microsoft Corporation
Redmond, WA
jacob.spoelstra@microsoft.com

## ABSTRACT

Acne is the 8th most common skin disorder in the world [8]. Assessment of acne severity is typically done by dermatologists in a clinical environment. We worked with Nestlé SHIELD (Skin Health, Innovation, Education, and Longevity Development, NSH) to develop a deep learning model that is able to assess acne severity from selfie images as accurately as dermatologists. The model was deployed as a mobile application, providing patients an easy way to assess and track the progress of their acne treatment.

NSH acquired 4,700 selfie images for this study and recruited 11 internal dermatologists to label them in five categories: From 1 (Clear) to 5 (Severe). While convolutional neural networks (CNN) have become the leading approach for computer vision, they typically require much larger training sets. This is exacerbated by the spatial sensitivity of CNN models since acne severity is judged on localized skin conditions. In addition, we had to address frequent poor image quality.

We implemented a transfer learning approach by extracting image features using a ResNet 152 pre-trained model, then adding and training a fully connected layer to learn the target severity level from labeled images.

In order to minimize irrelevant background, we used OpenCV models to find facial landmarks which were then used to extract key skin patches from the selfie images. To address the spatial sensitivity of CNN models, we introduce a new image rolling augmentation approach. Conceptually, it causes acne lesions to appear in more locations in the training images and improved the generalization of the CNN model on test images.

Evaluated on 230 test images using RMSE with respect to a consensus among experts, our model outperformed more than half of the human panel.

To our knowledge, this is the first deep learning-based solution for acne assessment using selfie images.

## 1 Introduction

Acne, medically known as acne vulgaris, occurs when pores become clogged with dead skin cells and oil. This creates blackheads, whiteheads and, as inflammation worsens, red pimples [9]. Traditionally, acne severity assessment is made by the dermatologists in a clinical environment. Prescribed treatments are carried out by dermatologists, or over-the-counter skin care products are recommended based on the severity [7].

Because the disease is so common, demand from the acne patients who would like to have their acne severity assessed professionally on a regular basis outstrips the availability of dermatologists to assess the acne severity. It is estimated that acne patients must wait for an average of over 32 days for an appointment with their dermatologist [1]. This presents a big challenge and frustration for acne patients since it delays guidance on diet, life style and skin care products.

To fill this gap, NSH collaborated with Microsoft to develop a consumer mobile app for acne assessment. There are two main functions of the mobile app: 1) accurately assess the acne severity of users based on uploaded selfie images, and 2) recommend treatment plans appropriate to the specific level of severity, factoring in demographic information such as gender, age, skin type, etc. A brief demo of the mobile app is available on YouTube[1] starting at 13'50''.

While the app is not intended to replace a visit to dermatologist for clinical assessment, it allows NSH to actively engage with their customers, make dermatological expertise user-friendly and accessible to customers, and address their customers' needs promptly. This application significantly shortens the feedback loop for their products with real-time customer data, thus enabling the company to proactively design new or improve existing products.

Previous work on biomedical image classification has focused on using high resolution or close-view images instead of selfie images. Esteva et al. [4] trained CNN models using 129,450 high-resolution and focused images to classify images as benign and malignant skin cancer. Choi et al. [3] built random forest transfer learning model using pretrained VGG-19 model as feature extractor to classify 10

---

[1] https://youtu.be/7tgJsms0viI

classes of retinal images. The training dataset included 397 close-view and focused images.

None of the previous work on acne assessment has been deployed into a real mobile app and aligned with recommending skin care products. NSH identified a need for a tool which would allow consumers to safely and accurately self-assess, treat and track the healing process.

The main contributions of this work are summarized as follows:

(1) We describe the first mobile application for acne assessment that has accuracy on the level of a dermatologist using just cell phone selfie images instead of clinical high-resolution images.
(2) We propose a novel image augmentation approach for facial images which addresses the spatial sensitivity problem of CNN models on small training data. It significantly improves the generalization of the model on test images.
(3) We build a real-world skin management mobile application facilitating the whole treatment cycle comprising dermatologists, users, and skin care products.

## 2 Approach

### 2.1 Training and Testing Images

NSH acquired 4,700 selfie images for this study. Among those, 230 were selected by a NSH senior dermatologist to serve as a 'golden set' for model evaluation, and the remaining 4,470 images were used as training images. The training images were randomly split into 11 groups and assigned to 11 dermatologists to label [5]. The "golden set" images were assigned to all 11 of the dermatologists, allowing us to calculate an average score.

There were a significant number of low-quality training images due to over- or under-exposure, or very low resolution. We decided to discard those, leaving about 1,000 training images.

Although the acne severity level is categorical, it is reasonable to use ordinary numerical values for them: 0-Not Acne, 1-Clear, 2-Almost Clear, 3-Mild, 4-Moderate, and 5-Severe.

Figure. 1 shows the distribution of the numbers of images that fell into different severity categories. Since the category 0-Not Acne did not have enough examples in both training and testing sets, we excluded it from the modeling and evaluation. It also shows that the image classes are imbalanced. Class 3-Mild dominates the acne severity levels in both training and testing images.

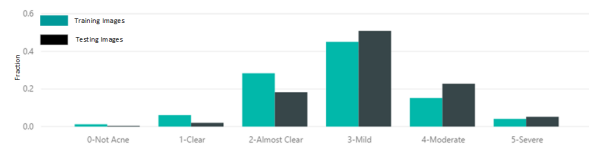

**Figure 1. Distribution of Images in Categories of Acne Severity**

### 2.2 Image Processing

Following a common strategy to address the problem of limited data, we applied the following image processing steps to augment the set of training images. The objective of these steps was to minimize noise introduced by the background, over- or under-exposure, low resolution, and to make the trained model spatially robust. This turned out to improve the generalization performance on the testing images. The remainder of this section describes the various steps in this pipeline.

*2.2.1 Extracted skin patches from facial skins.* Although acne may happen anywhere on a face, the dominant areas of acne are forehead, cheeks and chin. In this step, we extracted skin patches from the forehead, both cheeks, and chin using the OpenFace face recognition library facial landmark model. It has proved to achieve near-human accuracy in a benchmark dataset [2, 6].

We propose a coupling method for skin patch extraction for both frontal face images and side view images. The facial landmark model is the first step. If no face was detected in the first step, we employed the OpenCV One Eye model to detect the location of the single eye. The OpenCV model is an object detection model using a Haar feature-based cascade classifier [10]. Based on the eye location, we inferred the regions of the forehead, cheeks and chin skin patches. So, from each face image we extracted 2-4 skin patches depending on the image view angle. The same overall acne severity label assigned by the dermatologists was assigned to all skin patches from the same image.

Figure 2 shows the results of the skin patches from landmark facial detection. Each patch was labeled with the name to differentiate between the skin patches.

*2.2.2 Rolling skin patches.* CNN models are spatially sensitive. It means that if a feature on a test image appears on a new location where the same feature had never been seen before on the training images, the CNN model cannot recognize it.

This brings a significant challenge. First, the severity of acne does not depend on where the acne lesion is. Instead, it is mostly determined by the severity of each acne lesion, and how many there are. Second, the number of training images is small, and the acne lesions only appear in a limited number of locations. Therefore, the locations of acne lesion on the testing images are very likely to be new compared to the training images. The consequence is that the CNN models do not generalize well on the test images.

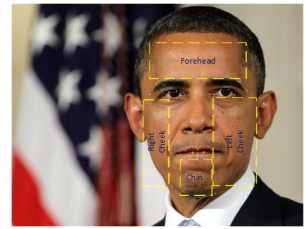 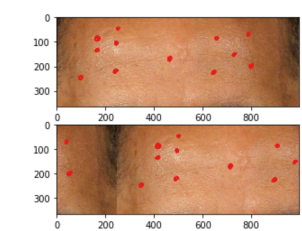

**Figure 2: Facial Skin Patches**    **Figure 3: Skin Patch Rolling**

To mitigate this problem, we rolled each skin patch for a certain number of pixels. Forehead image patches are rolled from right to left as shown in Figure 3. Cheek and chin image patches are rolled bottom to top. The rolling step size is described in Equation (1):

$$roll\_size = \text{int}\left(\frac{X}{N+1}\right) \quad (1)$$

Where X represents number of pixels in the rolling direction. $N$ is the number of rolling times.

Parameter $N$ was chosen to make the skin patches after rolling among all 5 classes nearly balanced. Patches of Mild images were rolled 2 times in order to help the CNN model generalize on the mild-class as well.

The Python code below achieves the horizontal rolling of an image:

```
roll_img[:,0:w-roll_size,:]  = raw_img[:,roll_size:w,:]
roll_img[:,w-roll_size:w,:] = raw_img[:,0:roll_size,:]
```

## 2.3 Model Development

*2.3.1 Converting Classification to Regression Problems.* One of the challenges in this work was that image labels from the dermatologists were noisy. We noticed that there were multiple identical (or close-to-identical) images in the training image set which were labeled differently by different dermatologists. Figure 4 shows the wide distribution of labels given by different dermatologists on several example testing images.

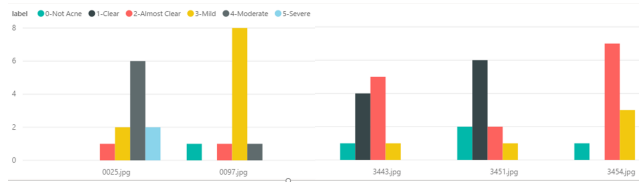

**Figure 4. Wide Distribution of Labels on Testing Images**

Given the ordinal nature of the labels, a regression model instead of classification was built. A higher numerical value represents more severe acne lesions.

*2.3.2 Transfer Learning.* Transfer learning utilizes the knowledge from one area to solve the problems in another area. We used a pretrained model to extract features from the lower layers, then trained a customized neural network model to solve our problem.

We used a pre-trained deep learning model to extract features from the training image skin patches. Considering both performance and computational efficiency, experiments showed ResNet 152 provided by CNTK performed best. We then trained a 3-layer fully-connected regression neural network model (with 1024, 512, and 256 hidden neurons) on these features to make the entire deep learning model specific to the acne severity domain.

## 2.4 Performance Metrics

NSH set as a success criterion for this project that the machine learning system to be at least as accurate on the test images as the "worst" among the 11 dermatologists.

As defined in equation (2), Root Mean Squared Error (RMSE) was chosen as the performance metric, with ground truth for each test image calculated as the average of the 11 numerical labels from the panel of dermatologists.

$$RMSE_i = \sqrt[2]{\frac{\sum_{k=1}^{N}(y_{ik}-\bar{y}_k)^2}{N}} \quad (2)$$

where i denotes the ith dermatologist, i = 1, 2 …, 11; N is the number of test (golden set) images; k = 1, 2, …, N; $y_{ik}$ is the label provided by dermatologist i on each k; $\bar{y}_k$ represents the average score from all dermatologists on image k.

To evaluate the model, it was used to individually score all of the skin patches extracted from each test image. The final score of the test image was then calculated as the average of the scores of all skin patches.

## 3 Experiments

### 3.1 Performance of Dermatologists

The RMSE for each dermatologist is shown in Table 1. The worst RMSE was 0.517, which set the bar for our model.

Table 1. RMSE of each Dermatologist

| Derma ID | 1 | 2 | 3 | 4 | 5 | 6 |
|---|---|---|---|---|---|---|
| RMSE | **0.517** | 0.508 | 0.500 | 0.495 | 0.490 | 0.484 |
| Derma ID | 7 | 8 | 9 | 10 | 11 | |
| RMSE | 0.454 | 0.450 | 0.402 | 0.400 | 0.388 | |

### 3.2 Performance of CNN-based Regression Model

If we simply use the global mean to predict the severity of the test images, we obtain a RMSE of 0.78, reflecting the relatively narrow distribution of severity levels shown in Figure 1. We call this the baseline model. Without skin patch rolling on the training images, the trained model achieved a RMSE of 0.72 on the test images. After we applied the skin patch rolling data augmentation, the RMSE on the testing images dropped to 0.482, which was better than the median RMSE of the 11 dermatologists.

Since the dermatologists were limited to discrete classes, we also evaluated our model as a classifier. We rounded the regression results (numerical values) to classification categories by using [1.5, 2.5, 3.5, 4.5] as category boundaries. The resulting classes were used to generate a confusion matrix as shown in Figure 5.

Figure 6 shows the scatter plot of the consensus acne severity levels (ground truth) vs. predicted severity levels (average of severity levels of skin patches) of the golden set images. There is a strong positive correlation (0.755) between the predicted and actual target values.

The CNN-based regression model performs well for images of mild acne (recall 82%), but the ability to differentiate Almost Clear (2) from Mild (3); Moderate (4) from Mild (3), and Severe (5) from Moderate (4), was still unsatisfying. We postulate that this is due to the noise in the training labels, where some identical images are labeled differently by different dermatologists.

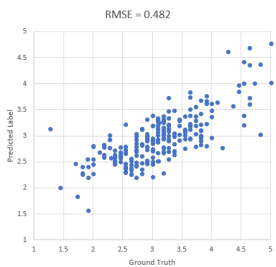

**Figure 5: Confusion Matrix of Discretize Severity Levels**

**Figure 6: Real vs. Predicted Severity**

### 3.3 Operationalization

A key requirement of this real-world project was to deploy the system as a practical consumer mobile app. We operationalized the trained CNN model, together with the image augmentation steps as a Python Flask web service API using Azure Container Service (ACS) and Azure Kubernetes Service (AKS). This exposed a web service API which could be invoked from the mobile app, sending the selfie image as payload and returning an acne severity score.

## 4 Conclusion and Future Recommendations

We described a practical approach to develop a CNN-based transfer learning regression model to assess the severity level of acne lesions from selfie images. The model was able to perform at the level of a trained dermatologist, allowing NSH to deploy a consumer app to assist in the treatment of this disease.

We proposed the coupling approach of facial landmark model and the OneEye OpenCV model to extract skin patches from different sectors of the face to eliminate background noise. Image rolling as an innovative data augmentation approach was critical to address the limited availability of labeled training data. Our results demonstrate that transfer learning regression with appropriate data augmentation is an effective method to train a domain-specific model despite a small sample training set.

Source code for the models described here can be found on [Github](https://github.com/Microsoft/nestle-acne-assessment)[2]. The repository includes the image processing and augmentation, deep learning models, and scripts for operationalizing the entire scoring pipeline as a web service API on AKS. The code can be generalized to other facial skin related computer vision problems besides this work.

There are a number of avenues for further improving the machine learning model:

(1) Label each skin patch separately: We used the label of the entire face image as the label of each skin patch. This step introduces additional noise to the labels of the training skin patches as the severity of acne lesion on different skin patches are different. Future work would require dermatologists to label the skin patches separately.

(2) Obtain more labeled images: The model could potentially be improved if more labeled images are available. After the mobile app is launched publicly, NSH will be able to collect more selfie images from customers. The dermatologists will be looped in labeling new images, which can be used to retrain the models on a regular basis.

(3) Use meta data of images to enhance the quality of training images: The model can also be improved if the selfie images for model training can be corrected for bad quality due to over- or under-exposure. This would be possible if meta-data for the images were captured. The mobile app has enabled the function of recording the meta-data of images. This will make more images with good quality available for model training.

NSH and Microsoft are working together to make this application globally available to the millions of people who are affected by acne. The app will enable an individualized experience for each customer with instantaneous analysis of selfie images and personalized guidance for treatment and management using interactive coaching.

### ACKNOWLEDGMENTS
The authors acknowledge our collaborators on this project from both NSH and Microsoft. In particular, we thank our project manager, Vikram Paramasivan for initiating the project and guiding it to completion.

---
[2] https://github.com/Microsoft/nestle-acne-assessment